\documentclass{ieeeaccess}
\usepackage{amsmath,amssymb,amsfonts}
\usepackage{algorithmic}
\usepackage{graphicx}
\usepackage{textcomp}
\usepackage{soul}
\def\BibTeX{{\rm B\kern-.05em{\sc i\kern-.025em b}\kern-.08em
    T\kern-.1667em\lower.7ex\hbox{E}\kern-.125emX}}

\usepackage{hyperref}
\newcommand{\ml}{ML}

\newcommand{\mlmi}{\ml{}MI}

\usepackage[square,numbers,compress]{natbib}
\begin{document}
\history{
Date of publication April 09, 2024, date of current version April 09, 2024.
}
\doi{10.1109/ACCESS.2024.3387702}


\title{A Framework for Interpretability \\in Machine Learning for Medical Imaging}
\author{
\uppercase{Alan Q. Wang}\authorrefmark{1,2},
\uppercase{Batuhan K. Karaman}\authorrefmark{1,2}, 
\uppercase{Heejong Kim}\authorrefmark{2}, 
\uppercase{Jacob Rosenthal}\authorrefmark{2,3}, 
\uppercase{Rachit Saluja}\authorrefmark{1,2}, 
\uppercase{Sean I. Young}\authorrefmark{4,5}, 
and \uppercase{Mert R. Sabuncu}\authorrefmark{1,2} 
}

\address[1]{School of Electrical and Computer Engineering, Cornell University and Cornell Tech, New York, NY 10044, USA}
\address[2]{Department of Radiology, Weill Cornell Medical School, New York, NY 10065, USA}
\address[3]{Weill Cornell/Rockefeller/Sloan Kettering Tri-Institutional MD-PhD Program, New York, NY 10065, USA}
\address[4]{Martinos Center for Biomedical Imaging, Harvard Medical School, Boston, MA 02129, USA}
\address[5]{Computer Science and Artificial Intelligence Laboratory, MIT, Cambridge, MA 02139, USA}
\tfootnote{Funding for this project was in part provided by the NIH grants R01AG053949 and T32GM007739, and the NSF CAREER 1748377 grant.}

\markboth
{Wang \headeretal: Preparation of Papers for IEEE TRANSACTIONS and JOURNALS}
{Wang \headeretal: Preparation of Papers for IEEE TRANSACTIONS and JOURNALS}

\corresp{Corresponding author: Alan Q. Wang (email: aw847@cornell.edu).}

\begin{abstract}
Interpretability for machine learning models in medical imaging (\mlmi{}) is an important direction of research.
However, there is a general sense of murkiness in what interpretability means. 
Why does the need for interpretability in \mlmi{} arise? 
What goals does one actually seek to address when interpretability is needed? 
To answer these questions, we identify a need to formalize the goals and elements of interpretability in \mlmi{}. 
By reasoning about real-world tasks and goals common in both medical image analysis and its intersection with machine learning, we identify five core elements of interpretability: localization, visual recognizability, physical attribution, model transparency, and actionability.
From this, we arrive at a framework for interpretability in \mlmi{}, which serves as a step-by-step guide to approaching interpretability in this context.
Overall, this paper formalizes interpretability needs in the context of medical imaging, and our applied perspective clarifies concrete MLMI-specific goals and considerations in order to guide method design and improve real-world usage. 
Our goal is to provide practical and didactic information for model designers and practitioners, inspire developers of models in the medical imaging field to reason more deeply about what interpretability is achieving, and suggest future directions of interpretability research.
\end{abstract}

\begin{keywords}
Interpretability, explainability, medical imaging, machine learning. 
\end{keywords}

\titlepgskip=-21pt

\maketitle

\begin{figure*}[t]
\centering
\includegraphics[width=\textwidth]{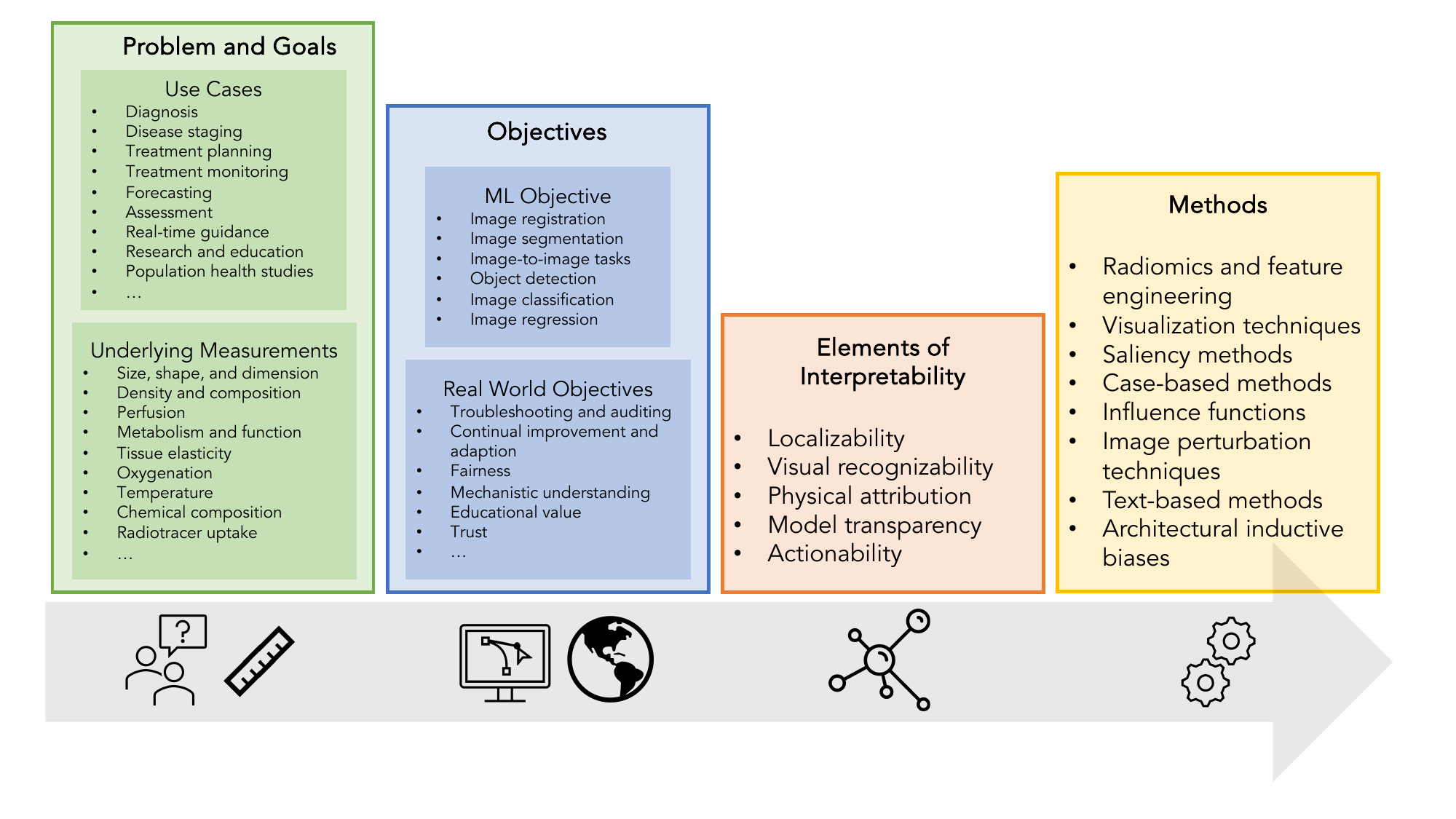}
   \caption{A framework for interpretability in \mlmi{}.  
    }
\label{fig:framework}
\end{figure*}

\section{Introduction}
\label{sec:introduction}
Machine learning (ML) has seen remarkable advancement in recent years. 
ML's intersection with medical imaging (which we abbreviate as \mlmi{}) is amongst the most promising, offering potential advances to quality of patient care~\citep{dembrower023aibreast,erickson2017mlmi,shen2017mlmi}.
However, the most performant machine learning models like those in computer vision and deep learning are generally regarded as black boxes -- they output predictions without revealing to human users how they arrived at those predictions.
As such, there has been a surge of papers calling for and proposing methods that make their decision-making interrogable, understandable, or explainable by users.
This subfield has gone by names like ``interpretable," ``explainable," and ``transparent" ML.
There is a clear need for such methods~\cite{rajpurkar2023current,reyes2020oninterpretability}, and the rising interest in this field is a reflection of the safety-critical, high-stakes setting in which medical imaging applications are deployed.

However, in the medical imaging field, there is a general sense of murkiness in how these words are or can be used. 
Many works claim a certain approach to be ``non-interpretable'' and thus sub-optimal, while others claim to increase interpretability, without a formalism that we can commonly agree on.
Our motivation in this paper is to introduce a formal framework to use for considering, motivating, studying, validating, and discussing interpretability in MLMI.
Central to our work is answering these questions:
\begin{enumerate}
\item 
Why does the need for interpretability in \mlmi{} arise?
\item 
What does one actually seek when interpretability is needed?
That is, what are the underlying \textit{elements} of interpretability in \mlmi{}?
\item 
Informed by the above two points, how can we formalize them into a framework for interpretability in \mlmi{}?
\end{enumerate}


This paper attempts to address these three points.
In particular, within the domain of medical imaging, we pinpoint elements from a purely applications-based perspective, starting from enumerating common tasks in medical image analysis and its intersection with ML, reasoning about their real world goals and considerations, and distilling from these five underlying elements: \textbf{localizability}, \textbf{visual recognizability}, \textbf{physical attribution}, \textbf{model transparency}, and \textbf{actionability}.
This culminates in a framework for interpretability in \mlmi{}, which serves as a step-by-step guide to approaching interpretability in this context.

We believe this paper can act as a guide for users and designers of interpretability in the medical imaging field by clarifying concepts in this context, suggesting future directions of research, and reasoning more deeply about how to evaluate the efficacy of their proposed methods.
In addition, we hope that our work can provide concrete and practical utility for clinicians and researchers who may benefit from these models in their own workflows.

The paper is organized as follows.
In Section~\ref{sec:related-works}, we overview works similar to ours from both general machine learning and medical imaging perspectives, and identify the gap which our work aims to fill.
In Section~\ref{sec:media}, we broadly describe the goals and measurements which are sought in medical image analysis.
In Section~\ref{sec:tasks}, we enumerate real-world tasks which comprise medical image analysis and which ML methods are employed to solve, as well as list possible real-world objectives which might be sought in the context of medical imaging and which cannot be explicitly optimized for.
In Section~\ref{sec:elements}, we formally describe the elements of interpretable \mlmi{} distilled from these real world tasks and goals.
In Section~\ref{sec:methods}, we connect these elements to existing literature by reframing existing methods.
These sections culminate in Section~\ref{sec:framework}, which provides a step-by-step guide to practitioners on approaching interpretability for \mlmi{}.
Finally, in Section~\ref{sec:discussion}, we discuss further implications, including 
possible future directions of exploration and limitations of interpretability as a whole.

To summarize, our contributions in this work are as follows:
\begin{itemize}
    \item We introduce a framework for interpretability in MLMI, starting from goals of medical image analysis and motivated by the need for various real-world considerations in this context, including trustworthiness, continual adaptation, and fairness.
    \item In doing so, we identify key elements of interpretability in MLMI and tie them to previously-proposed interpretability methods, in an effort to enable practical utility for clinicians and researchers.
    \item We discuss implications, limitations, and potential future exploration directions, aiming to foster deeper understanding and evaluation of interpretability methods in MLMI.
\end{itemize}

\section{Related Works}
\label{sec:related-works}
Interpretability in machine learning has been discussed and argued for by many articles~\cite{rudin2019stop,lipton2017doctor,kundu2021aimustbe}.
Several papers in interpretable machine learning attempt to define desiderata~\cite{lipton2017mythos,doshivelez2017rigorous,murdoch2019definitions}.
Fundamentally, these surveys identify the need for interpretability arising from a mismatch~\cite{lipton2017mythos} or incompleteness~\cite{doshivelez2017rigorous} between the objectives optimized during training (i.e. test set predictive performance) and the real world costs and benefits in a deployment setting.
Lipton enumerates several desiderata for interpretability, including trust, causality, informativeness, transferability, and fair and ethical decision-making~\cite{lipton2017mythos}.
Murdoch et al. view interpretability from the context of the life-cycle of models, including the relationship of the model with the practitioner and the audience~\cite{murdoch2019definitions}.
From their vantage point of general machine learning, these above surveys strive to cover all aspects of interpretability in this domain.
While broad, they are necessarily vague in order to encompass all facets of ML, sacrificing practical utility in the process.

Our work follows the guidelines of Rudin~\cite{rudin2019stop}, who argues that interpretability should be defined in a domain-specific manner.
However, even in the narrower context of \mlmi{}, there is not a consensus in the literature on what precisely defines interpretability.
Many review papers survey the landscape of interpretability methods available to medical imaging practitioners like clinicians and researchers ~\cite{salahuddin2022transparency,huff2021interpretation,vandervalden2022xai}.
For example, Huff et al. categorize models according to whether they seek to understand model structure and function or whether they seek to understand model output~\cite{huff2021interpretation}.
van der Valden et al. propose a categorization based on model-based vs. post-hoc, model-specific vs. model-agnostic, and global vs. local methods~\cite{vandervalden2022xai}. 
While comprehensively categorizing available methodologies, these surveys are not focused on obtaining or distilling from these methods any underlying elements.

Some works go beyond categorizing methodologies and propose higher-level frameworks for interpretability in this narrower context.
Schuhmacher et al. propose a framework for falsifiable explanations in medical imaging contexts~\cite{schuhmacher2022falsifiable}.
Their framework requires that any interpretability method connect to the physical reality of the sample from which the data originate, and also require an explanation to be a falsifiable hypothesis. 
Their framework can be seen as a middle ground between the inductive nature of machine learning training and the hypothetico-deductive nature of empirical research.
Our framework can be seen as an elaboration upon this framework with additional elements which we believe render it more comprehensive.

Common to all aforementioned works is the fact that any core properties which are gleaned from the literature are distilled from previously proposed papers and methods.
Typically, these works draw a distinction between models that are inherently interpretable by design, and models that are interpretable post-hoc, possibly by a separate trained model~\cite{rudin2019stop,murdoch2019definitions,lipton2017mythos}.
We believe such a meta-analytical approach runs counter to the applied nature of the field.
Instead, in this work, we pinpoint elements by first considering real-world tasks and considerations surrounding medical image analysis and its intersection with ML.
We distill elements of interpretability from this applied perspective, without being influenced by current literature. 
An approach which starts from available methodologies is liable to restrict itself to elements which are currently feasible, instead of those which are truly desired. 

\begin{figure*}[t]
\centering
\includegraphics[width=\textwidth]{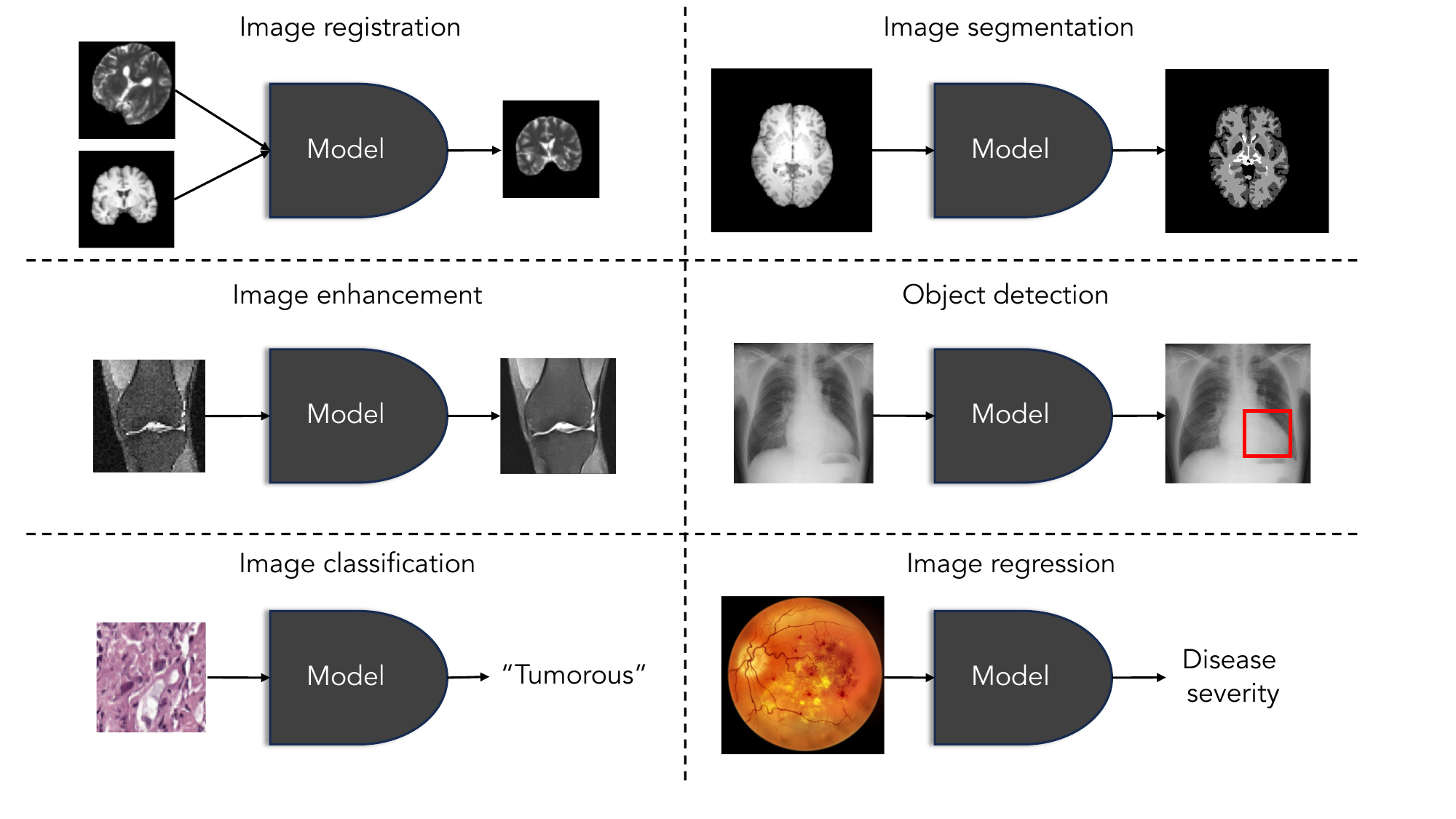}
   \caption{Common tasks in \mlmi{}.  
   Tasks are primarily characterized by the structures of their input features and output predictions.
    In MLMI, inputs are images or features derived from images, sometimes combined with meta-data such as patient information. 
    The structure of output predictions are determined by the task.
    }
\label{fig:tasks}
\end{figure*}

\begin{figure*}[t]
\centering
\includegraphics[width=\textwidth]{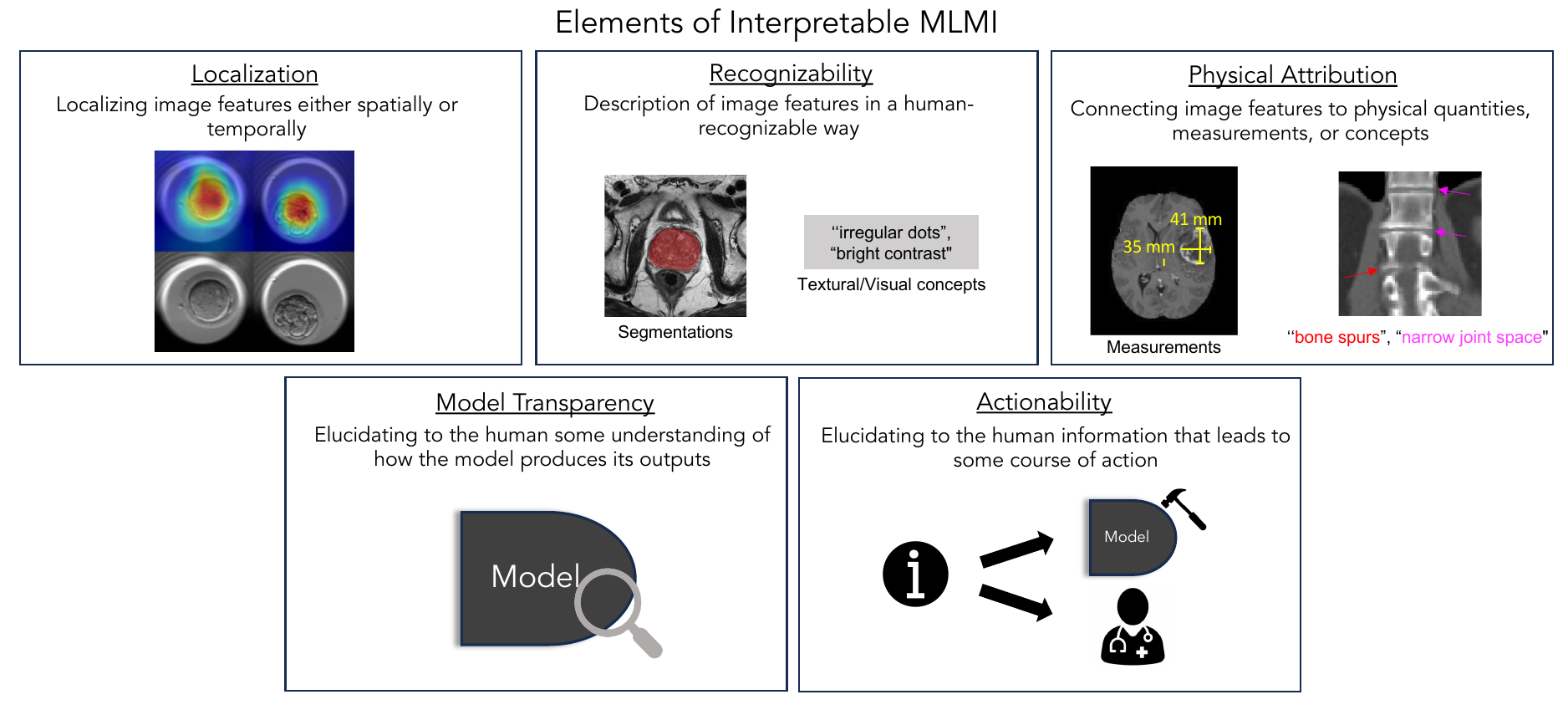}
   \caption{Graphical overview of the elements of interpretable \mlmi{}.
   }
\label{fig:flow}
\end{figure*}

\section{Goals of Medical Image Analysis}
\label{sec:media}
Medical image analysis is a multifaceted discipline, comprising many different imaging modalities, anatomies, and end goals.
To provide a solid foundation upon which to clarify goals and elements for interpretable \mlmi{}, we start by enumerating the use cases of medical image analysis, as well as the underlying measurements which are derived from medical imaging.

\subsection{Use Cases}
\label{sec:use-cases}
Here, we enumerate several real-world use cases of medical image analysis.
The ultimate goals for these use cases could be clinical or scientific in nature.
On the clinical side, some example goals might include pathology detection and characterization such as in bone fracture assessment, screening high risk groups such as with lung imaging, and assessing treatment outcomes such as with molecular imaging.
On the scientific side, example use cases include cell biology, neuroscience, and drug development.

\subsubsection{Diagnosis} 
Medical imaging can help in identifying the nature of an illness~\cite{FernándezMontenegro2020alzheimers,Nooreldeen2021lungcancer,Bussani2020cardiactumors}, which may require revealing abnormalities such as tumors, fractures, infections, or organ damage that may not be apparent through physical examinations alone.

\subsubsection{Disease Staging} 
The stage and severity of diseases can be facilitated by medical imaging~\cite{conklin1984disease}. For example, medical images in oncology can show the size and spread of tumors, enabling identification of the cancer severity and furthermore enable treatment planning.

\subsubsection{Treatment Planning} 
Treatment planning is concerned with tailoring the application of available treatment resources to each patient's individual goals and needs~\cite{johnson2013treatmentplanning,noel2009treatmentplanning}.
Common examples include planning surgical procedures and radiation therapy. 
With medical imaging, surgeons can visualize the precise location and extent of abnormalities, helping them make informed decisions about the best approach to treatment.

\subsubsection{Treatment Monitoring} 
Similar to treatment planning, treatment monitoring seeks to assess the effectiveness of treatments over time~\cite{aronson2005monitoring}. Medical imaging can monitor changes in the size and characteristics of tumors or other disease markers, which can indicate how well a treatment is working.

\subsubsection{Forecasting} 
Also known as prognostication, forecasting is concerned with predicting the likely course of a disease and its potential outcomes~\cite{chu2019prognostication,hui2019prognostication}. 
As a prerequisite, early detection of diseases can be facilitated through routine screening using imaging techniques like mammography or computed tomography (CT) angiography.

\subsubsection{Assessment} 
Medical imaging is useful for assessing function of different parts of the body using imaging techniques such as functional magnetic resonance imaging (fMRI) and positron emission tomography (PET) scans. This is valuable for understanding brain activity, blood flow, and metabolic processes. Medical images also offer ways to interrogate structural health, such as in x-rays and bone density scans (DXA) used to evaluate bone health, detect fractures, and assess the risk of osteoporosis.

\subsubsection{Real-Time Guidance} 
Real-time guidance for minimally invasive procedures~\cite{antico2019guidance,stull2019robotic}, such as biopsies, catheter insertions, and needle aspirations helps to ensure precision and minimize risks.

\subsubsection{Research and Education} 
Medical imaging data is invaluable for clinicians and scientists for purposes of research and education. For example, histopathology imaging is an important tool for researchers studying the human body~\cite{srinidhi2021deep}, allowing them to investigate diseases and develop new treatments. 
Additionally, medical imaging is an important tool for educating medical students and other healthcare professionals.

\subsubsection{Population Health Studies} 
Population health studies utilize aggregated medical imaging data to identify trends, risk factors, and patterns of disease within communities~\cite{gourevitch2019populationhealth,shahzad2019population}.

\subsection{Underlying Measurements}
\label{sec:measurements}
Fulfilling these use cases may require various physical measurements which are uncovered through medical imaging.
In this section, we enumerate several types of underlying signals that are measured in and derived from various imaging modalities.
In subsequent sections, we consider how connecting and referencing to such measurements are, amongst other factors, central to interpretable \mlmi{}. 

\subsubsection{Size, Shape and Dimension} 
Medical imaging can precisely measure the size, shape, and dimensions of organs, tumors, lesions, and other anatomical structures~\cite{shapeMI2020,singh2023topological,nguyen2020application}. Techniques like ultrasound, X-ray, CT, and magnetic resonance imaging (MRI) are commonly used for this purpose.

\subsubsection{Density and Composition} 
Imaging techniques such as X-ray and CT scan provide information about tissue density and composition~\cite{mesbah2017density,yaffe2008density}. Different tissues have varying X-ray absorption characteristics, allowing the differentiation of bone, soft tissue, and air, for example.

\subsubsection{Perfusion} 
Perfusion imaging measures blood flow to tissues and organs~\cite{demeestere2020perfusion,kamphuis2020perfusion}. Techniques like dynamic contrast-enhanced MRI and nuclear medicine scans (e.g. PET) can assess tissue perfusion, helping diagnose conditions like stroke or assess the viability of transplanted organs. 

\subsubsection{Metabolism and Function} 
PET and fMRI are imaging techniques that can probe metabolic activity and functional processes in the body~\cite{judge2020metabolism,glover2011overview}. They are used to study brain function, detect cancer, and assess organ function.

\subsubsection{Tissue Elasticity} 
Elastography is a specialized ultrasound technique that measures tissue elasticity or stiffness~\cite{sigrist2017elastography,sarvazyan2011elastography}. It is valuable in assessing the stiffness of liver tissue in conditions like cirrhosis and for breast cancer diagnosis.


\subsubsection{Oxygenation} 
Near-infrared spectroscopy (NIRS) and functional MRI can measure tissue oxygenation and oxygen levels in the blood, aiding in the assessment of brain function and detecting oxygenation problems in tissues~\cite{dantzker1993oxygenation,glover2011overview}.

\subsubsection{Temperature} 
Infrared thermography uses thermal imaging to measure the temperature of the skin’s surface~\cite{sarawade2018thermography}. It has applications in detecting inflammation and circulatory problems~\cite{lahiri2012thermography}.


\subsubsection{Chemical Composition} 
Magnetic Resonance Spectroscopy (MRS) is a specialized MRI technique that can provide information about the chemical composition of tissues, including the presence of specific metabolites~\cite{tognarelli2015mrs}.

\subsubsection{Radiotracer Uptake} 
PET scans use radiotracers labeled with radioactive isotopes to measure the uptake of specific substances in tissues, such as glucose or neurotransmitters~\cite{OSullivan2015bonemetastasis,fortunati2022radiotracers}.

\section{MLMI Objectives}
\label{sec:tasks}

\subsection{Primary ML Objectives}
\label{sec:ml-tasks}
As evidenced by the previous section, medical image analysis represents an array of use cases as well as imaging-derived measurements which are necessary to fulfill these use cases.
MLMI is therefore comprised of many different ML tasks.
In ML, tasks are primarily characterized by the structures of their input features and output predictions.
In MLMI, inputs are images or features derived from images, sometimes combined with meta-data such as patient information. 
The structure of output predictions are determined by the task.
In this section, we detail some of the most common tasks, and visualize them in Fig.~\ref{fig:tasks}.

\begin{itemize}
\item
\textit{Image registration}~\cite{zitova2003registration} seeks to align scans of the same anatomy. 
The output is typically a transformed image aligned to another fixed image.

\item
\textit{Image segmentation}~\cite{minaee2022segmentation} seeks to delineate different anatomical structures into discrete categories. The output is pixel-level class predictions.

\item
\textit{Image enhancement, reconstruction, or restoration} predict a clean version of a noisy image~\cite{ahishakiye2021reconstruction}.
The output is pixel-level scalar predictions.

\item
\textit{Object detection}~\cite{kaur2022survey,ganatra2021objectdetection} seeks to detect or isolate specific objects in the image, like organs or lesions. 
The outputs can be image coordinates (e.g. landmarks) or bounding boxes.

\item
\textit{Image (or patient)-level classification} seeks to categorize an image at the patient-level into one of several distinct categories (e.g.~diagnosis)~\cite{yadav2019classification,doi2007cad}.
The output is a single class label.

\item
\textit{Image (or patient)-level regression} seeks to associate with an image a real number that may represent some quantity (e.g.~disease severity or some physical measurement)~\cite{cohen2020regressionfmri}.
The output is a scalar real number.
\end{itemize}

The task and its corresponding structure of output prediction typically imply a prescribed objective which is optimized (e.g.~mean-squared-error or cross-entropy).
This objective is mathematically defined and thus can be optimized explicitly.
However, in the real world, we often have considerations that go beyond these objectives.
This is especially true in the context of medical imaging, where image-based tasks form one component of a larger clinical workflow, and where real world considerations such as safety, trust, fairness, and ease-of-use must be balanced alongside mathematical loss functions~\cite{varoquaux2022mlmi}.   
In the next section, we turn to these real world goals.

\subsection{Other Real-world Objectives}
\label{sec:goals}
In the previous section, we enumerated common tasks in medical image analysis and characterized them by the structure of their predictions; this structure lends itself to natural objectives that quantify performance and can be explicitly optimized (such as Dice overlap score for image segmentation or accuracy for classification).
However, often times in real world deployment, we have goals beyond this natural objective.
In this section, we enumerate several such (non-exhaustive) goals.

\subsubsection{Facilitating Troubleshooting and/or Auditing.}
Clinicians and researchers may require that models enable some way of facilitating troubleshooting and/or auditing.
This is especially necessary when models are performing poorly, or when the robustness of models are called into question.
As an example, Zech et al.~used activation maps to uncover models leveraging spurious features (e.g. ``shortcuts") in chest radiographs~\cite{zechVariableGeneralizationPerformance2018}.
In particular, activation maps showed that image artifacts (such as indicators of R and L sides) that were heavily correlated with the outcome of interest were driving the prediction.
Such a finding might lead to removing such artifacts or other changes to remove model dependence on this shortcut.

\subsubsection{Enabling Continual Improvement and Adaptation.}
Related but distinct from troubleshooting and auditing, clinicians and researchers may require that models continually improve and adapt over time.
Data is a reflection of the real world and is therefore dynamic and constantly evolving.
This can stem from availability of more data over time, but also may stem from shifts in the underlying statistics of the images due to changes in prevalence of diseases, changes in imaging protocols, or upgrades in measurement technologies~\cite{finlaysonClinicianDatasetShift2021}.
A model should enable users to identify and correct errors due to these changes.

\subsubsection{Ensuring Fairness.} 
Fairness seeks to guarantee that a model's performance is balanced or invariant across patient groups, and in particular that it does not disadvantage minority or otherwise protected groups~\cite{caton2020fairness}. 
A clinician or researcher should be able to ensure and validate fairness of a model by, for example, interrogating and comparing the model's performance across groups. 
As an example, Puyol-Anton et al. identified statistically significant differences in Dice scores between white population and minority ethnic groups for state-of-the-art models that segmented ventricles and the myocardium in cine cardiac magnetic resonance images~\cite{puyol2022fairness}. 
As another example in brain MRI segmentation, Ioannou et al. observed a statistically significant drop in Dice scores when models trained on data from white subjects were evaluated on black female subjects~\cite{ioannou2022study}. 
Finally, Abbasi-Sureshjani et al. showed that skin lesion classifiers exhibited large differences in performance between male and female patients even though their training data was relatively balanced~\cite{abbasi2020risk}. 

\subsubsection{Uncovering Scientific Insights and Advancing Mechanistic Understanding.}
An exciting value proposition is the possibility of uncovering or revealing novel scientific insights.
Science and medicine are often concerned with discovering and characterizing causal and/or mechanistic relationships.
These relationships can be characterized on physical, physiological, or biochemical grounds via known mechanisms, phenomena, or pathways~\cite{ross2021causal}.

As an example, a model trained to predict survival from H\&E histological sections for colorectal cancer was interrogated using a combination of model interpretability approaches and manual review by expert human pathologists, and a prognostic histological feature, tumor adipose feature (TAF), was identified~\cite{wulczynInterpretableSurvivalPrediction2021}. 
Human pathologists were then able to learn to independently identify the TAF feature as an effective prognostic marker~\cite{limperioPathologistValidationMachine2023}.
Models could play an important role in multiple parts of the scientific process, from discovery of new image-based patterns to improving human performance on critical tasks.

\subsubsection{Providing Educational Value.}
Clinical trainees often need to learn to analyze several kinds of images to extract key clinical imaging features to identify patterns and produce differential diagnosis corresponding to those patterns. 
However, their training and education may be hampered due to time constraints, limited training opportunities, or a lack of access to educational images. 
Machine learning based teaching platforms can augment their educational experience and provide key insights to improve them as clinicians.
~\cite{cheng2020artificial} presents an AI-augmented medical education platform to detect hip fractures by showcasing a heat-map to indicate locations of abnormality. They demonstrated that students who were trained by their system showed an improvement in understanding of the key features of a hip fracture in radiographs. Adaptive Radiology Interpretation and Education System (ARIES) is another platform presented by \cite{rudie2021brain} which uses ML models to extract clinical imaging features from various brain MRI sequences and present it to neuroradiology residents to perform differential diagnosis. 
They observed that the residents performed significantly better with ARIES than without. 

\subsubsection{Engendering Trust.}
Perhaps the least tangible but equally important auxiliary objective is engendering trust.
The emphasis on trust is most needed in clinical settings, where one of the major barriers to integrating ML models into clinical workflows is a lack of trust among clinicians; from this perspective, interpretability plays a role in MLMI as a means by which to improve trust among humans who are using these models~\cite{rajpurkar2023current,salahuddin2022transparency,shen2022trust,reyes2020oninterpretability}.
The concept of trust can further be decomposed into two components: trusting individual predictions produced by the model, and trusting the model itself as an entity for producing predictions~\cite{ribeiroWhyShouldTrust2016}.
Notably, this implies that explainability is principally valuable as an instrument to achieve other end goals, such as building trust or increasing usability of models, but does not necessarily have intrinsic value in and of itself~\cite{mccoyBelievingBlackBoxes2022}.

\section{Elements of Interpretable \mlmi{}}
\label{sec:elements}
In the previous section, we enumerated various ML tasks and several real world goals for \mlmi{} models, which in conjunction with the explicit objective, can be seen as constituting the combined end goals of the model.
From the preceding section, it is apparent that real world models require insights that go beyond a single, aggregated, and quantifiable performance metric which encodes the explicit objective.
The need for such insights brings about the need for interpretability~\cite{lipton2017mythos,doshivelez2017rigorous}. 

Yet, it is clear that not all these goals are sought after for every task, and furthermore that each one of these goals demands a different ``element" of interpretability.
That is, there is not one unique element of interpretability that satisfies all real world goals in all scenarios.
Thus, we seek to answer the question: what are the core underlying elements of interpretability in \mlmi{}?
In particular, we seek to distill these objectives to be as minimal as possible. 

We pinpoint five (possibly non-exhaustive) underlying elements: \textbf{localizability}, \textbf{visual recognizability}, \textbf{physical attribution}, \textbf{model transparency}, and \textbf{actionability}. 
The first three are concerned with characterizing the features of the image.
The fourth element, model transparency, is concerned with the model itself.
The fifth element, actionability, is concerned with both the model and influence on the human user.
These elements are further detailed below and depicted in Fig.~\ref{fig:flow}.

\subsection{Localizability}
\textit{Where} are the features, either spatially or temporally, that are driving the prediction?
Medical images are high-dimensional spatio-temporal data that simultaneously exhibit relevant and redundant information, at various scales and locations.
\textbf{Localizability} of features which are driving the prediction and conveying this location to the human is a central element of interpretability in \mlmi{}.

\subsection{Visual Recognizability}
\textit{What} are the visual features that drive the prediction?
We refer to a model which exhibits \textbf{visual recognizability} as one that can provide a description of these image features in a human-recognizable way.
This might involve conveying to the user features related to brightness, certain pixel intensity patterns, or certain textural patterns.

\subsection{Physical Attribution}
This element seeks to reveal to the human salient features of the image which are attributed to some underlying \textbf{physical meaning}. 
This element connects image features to real-world entities of or related to the image formation process. 
These entities could be quantitative (measurements like those enumerated in Section~\ref{sec:measurements}) or conceptual (e.g. semantic or human-derived concepts).

\subsection{Model Transparency}
\textbf{Model transparency} elucidates to the human some degree of understanding of \textit{how the model produces its outputs}.
In contrast to the previous three properties, transparency is concerned with inputs, architectures, and/or algorithms and does not characterize image features. 

Prior works in the ML literature have discussed different notions of transparency, including simulatability, decomposability, algorithmic transparency, sparsity, modularity, and intelligibility~\cite{lipton2017mythos,murdoch2019definitions,alvarezmelis2018robust}.
These notions are applicable broadly to any ML model and are not special to medical image analysis.

By contrast, a more \mlmi{}-specific notion of transparency are approaches which incorporate information related to known entities or mechanisms with respect to the problem being solved or other domain-specific information.
Inductive biases can then be injected which constrain the model to operate within the bounds of that domain.
For example, the model designer may have domain knowledge about the data-generating and/or image-formation process~\cite{monga2020algorithm,arias2022physicsbased,castro2020causality}, or have prior knowledge of the mechanisms or anatomy that underlie the problem being studied~\cite{dalca2018anatomicalpriors,gaw2019integration,pisov2019incorporating}.
See Section~\ref{sec:inductivebiases} for further details.

\subsection{Actionability}
\textbf{Actionability} elucidates to the human information that \textit{leads to some course of action}.
The action may be some form of falsifiability or recourse~\cite{schuhmacher2022falsifiable}. 
We identify two types of actionability.

The first type refers to recourse with respect to the model. 
Here, actionability refers to interpretable information that makes apparent a solution which involves changing or editing some aspect of the model or algorithm. 
This is related to model transparency, in that models which are transparent can naturally lead to actionable recourse.

The second type refers to recourse with respect to the human user. 
Here, actionability refers to interpretable information which leads to actionable insights for the human user.
For example, making apparent certain pathologies in the image might correspond to prescribed protocols that the human should undertake.
As another example, an interpretation may indicate to the human to run a follow-up experiment to further interrogate the problem in question.

\section{Connecting Existing Methods to Elements} 
\label{sec:methods}

In this section, we discuss several popular interpretability methods and describe how they relate to our aforementioned elements.
For more comprehensive overviews of methodologies, we refer the reader to dedicated surveys~\cite{salahuddin2022transparency,huff2021interpretation,vandervalden2022xai}.

\subsection{Radiomics and Feature Engineering}
Radiomics and other clinically-derived features seek to reduce high dimensional features like shape, intensity and texture into features using image processing or clinically-informed techniques~\cite{lambin2012radiomics,gillies2016radiomics,lambin2017radiomics,rauschecker2020neuroradiologist,bera2022predicting}, and is particularly prevalent in oncological applications. 
Typical radiomics workflows require a segmentation step to first localize the region of interest (e.g. tumor), and thus may be seen to exhibit \textbf{localizability}.
Furthermore, since many engineered features are designed to capture intensity or textural information, they may be seen to exhibit some notion of \textbf{recognizability}.
Features focused on domain-specific and clinically-relevant features like shape, size, or orientation may be seen as exhibiting \textbf{physical attribution}.
Finally, because many features involve relatively simple mathematical expressions to compute, humans can typically contemplate the computation of these features and can thus be seen to exhibit \textbf{model transparency}.

\subsection{Visualization Techniques}
In this section, we detail several visualization techniques that seek to convey to humans the inner workings of a model via information that can be plotted.
Since these methods are all seeking to overcome the black-box nature of models, they can all be seen as exhibiting \textbf{model transparency}.

Feature importance techniques such as partial dependence plots (PDPs)~\cite{friedman_2001_greedy}, individual conditional expectation (ICE) plots~\cite{goldstein_2015_peeking}, and SHapley Additive exPlanations (SHAP)~\cite{lundberg2017unified,bang2021interpretable,zhang2023radiomics} have been used for analysing partial dependency of breast tumor malignancy on ultrasound image features~\cite{zhang_2010_partial}, and uncovering important features in predicting cognitive decline in Alzheimer's disease~\cite{karaman_2022_machine}. 

A class of techniques allows for analysis of high-dimensional latent space of deep networks.
Techniques like t-SNE~\cite{maaten2008tsne} and UMAP~\cite{mcinnes2020umap} project high-dimensional embeddings into a low-dimensional space that can be plotted (e.g. in 2 dimensions).
Doing so can roughly uncover how models organize datapoints in the latent space.

\subsection{Saliency Maps}
Saliency maps reveal salient features within an image such as lesions or changes over time in longitudinal studies by outputting heatmaps which can be overlaid on top of the image(s)~\cite{ zhang2017mdnet, bohle2021convolutional, sun2023inherently}. 
Popular techniques include CAM-based methods~\cite{zhou_2015_learning,selvaraju_2020_gradcam,draelos_2021_use,feng_2022_a,lee_2022_chexgat,draelos_2021_explainable,kim2023learning} and methods that incorporate attention mechanisms to make predictions~\cite{yan2019melanoma,yang2019guided,bohle2021convolutional,yin2023anatomically}.

Saliency maps may be seen to exhibit \textbf{localizability}, since they highlight salient features locally.

\subsection{Case-Based Methods and Explanation by Example} 
This class of methods present evidence for predictions for a query sample by identifying similar samples.
This is sometimes referred to as ``explanation by example''~\cite{lipton2017mythos,kolodner_2014_casebased}.
Prototypical networks \cite{snell2017prototypical,chen2019looks, kim2021xprotonet,wang2023nwhead} classify features or images (e.g.~chest X-ray scans) by finding their nearest class prototypes (e.g.~images of typical cardiomegaly, nodule, etc) in a learned embedding space. 
Lamy et al. propose a visual case-based reasoning method for classification and applied it to breast cancer~\cite{lamy_2019_explainable}.

Case-based methods may be seen to exhibit \textbf{model transparency}, since they can reveal to the user which images the model deems most (dis)similar to one another, which in turn reveals some understanding of its inner workings. 
In some cases, case-based methods may be paired with other techniques which exhibit other elements, like saliency maps for \textbf{localizability}~\cite{hu2022x}.

\subsection{Influence Functions} 
Influence functions quantify the most influential training images (either positive or negative) which lead to the prediction for a given sample~\cite{cook1982influence,koh2017influence,wang2023nwhead}.
For example, Wang et al. used influence functions in the context of liver tumor diagnosis in order to quantify the radiological features that were most influential for classifying a particular lesion~\cite{wang2019deep}.

Influence functions may be seen to exhibit \textbf{model transparency}, since they reveal the most ``helpful" or ``harmful" training images for a model to make a prediction for a given image.

\subsection{Image Perturbation Techniques} 
Image perturbation techniques directly perturb the input image. These perturbations can identify changes in intensity patterns and deformations that highlight underlying biophysical and biomechanical processes.
For example, counterfactuals~\cite{wachter_2017_counterfactual,nemirovsky_2021_countergan,zhou_2023_scgan} introduce the minimum amount of change to a given image to change its predicted class. 
Examples of counterfactual image generation for medical image analysis include chest \cite{sun2023inherently, sun2023right}, and knee  \cite{schutte2021using} X-ray scans for the prediction of pulmonary disorders, and severity of osteoarthritis, respectively, as well as brain \cite{baumgartner2018visual}, and breast \cite{zhou_2023_scgan} MRI for Alzheimer's disease and breast tumor analyses, respectively.

Other popular perturbation techniques include adversarial attacks~\cite{szegedy_2013_intriguing,du_2021_clinically,paul_2020_mitigating,goodfellow_2014_explaining,su_2019_one}, occlusion sensitivity~\cite{zeiler2013visualizing}, and local interpretable model-agnostic explanations (LIME)~\cite{ribeiroWhyShouldTrust2016}.

Perturbation techniques may be seen to exhibit \textbf{visual recognizability}, since they may alter features in meaningful ways that humans can recognize.
If these perturbations are meaningfully localized, they may also be seen to exhibit \textbf{localizability}.

\subsection{Text-Based Methods}
Many methods approach interpretability by incorporating human-derived and annotated concepts.
In \mlmi{}, these concepts could be clinically-relevant text or phrases, like ``sclerosis", ``bone spurs", or ``narrow joint space" in the context of arthritis grading.
Concept bottleneck models (CBMs)~\cite{koh2020concept, santurkar2021editing,patrício2023coherent} aim to decompose end-to-end prediction models into non-linear concept prediction and linear or logistic task prediction from the predicted concept values. 
Concept whitening (CW) models~\cite{chen2020concept} extend CBMs by additionally othogonalizing and rescaling the concept activation vectors during training for improved interpretability.
Applications of CBMs to medical image analysis include skin lesion diagnosis \cite{patricio2023coherent,yuksekgonul2022post}, pediatric appendicitis \cite{klimiene2022multiview}, and chest radiology \cite{chauhan2023interactive}.
Quantitative testing with concept activation vectors (TCAV)~\cite{kim_2018_interpretability} employs concept activation vectors to interpret the internal state of neural networks in terms of user-defined concepts; the authors apply it to a deep learning model that predicts diabetic retinopathy (DR) level from retinal fundus images. 

Other than concepts, text-based methods also may output justifications, reports, and captions. 
These methods typically provide long-form text along with the predictions.
Zhang et al. propose MDNet, which leverages a vision and language model to visualize attention as well as provide justifications of the predicted diagnosis~\cite{zhang2017mdnet}.
Lee et al. propose a visual word constraint model to improve the accuracy of a justification generator for diagnostic interpretations~\cite{lee2019generation}.

Concept-based methods may be seen to exhibit \textbf{physical attribution}, since concepts are typically human-annotated phrases that map to specific real-world characteristics associated with a diagnosis.
Additionally, text might point out textural or image-derived information (e.g. ``irregular dots" or ``bright contrast") which may be seen to exhibit \textbf{recognizability}.

\subsection{Architectural Inductive Biases} 
\label{sec:inductivebiases}
In medical imaging contexts, many techniques and methods exist which leverage domain knowledge to inject domain-specific inductive biases into model architectures.
These methods use information specific to the problem being solved, for example specific to image content (e.g. anatomy, geometry), or leveraging information about the data-generating process (e.g. a known forward model).
Since they directly alter architectures and algorithms used to solve the problem, models of this type may be seen to exhibit \textbf{model transparency}.

In reconstruction, unrolled architectures~\cite{monga2020algorithm} and physics-based models~\cite{arias2022physicsbased} incorporate knowledge of the underlying forward model, data-generating process, or physical constraints.
In classification, mechanistic models~\cite{baker2018mechanistic} and anatomical priors~\cite{dalca2018anatomicalpriors} incorporate knowledge of specific objects of interest into the architecture, for example brain midline shift~\cite{pisov2019incorporating}, models for tumor growth~\cite{gaw2019integration}, and cardiac image enhancement and segmentation~\cite{oktay2017anatomically}.
In registration, the relative homogeneity in geometry of medical images has been leveraged~\cite{he2023geometric}.
Another line of research seeks to apply causal inference to various contexts in \mlmi{}~\cite{causality2009pearl,castro2020causality,vlontzos2022review}.
Works in this vein inject prior information into the model by reasoning about the causal relationships in the data-generating process.

\section{A Framework for Interpretable MLMI}
\label{sec:framework}
In this section, we provide a guide to practitioners of \mlmi{} models for reasoning about interpretability.
This guide follows the ordering of sections in the paper, and we reference the corresponding section throughout.
A diagram is depicted in Fig.~\ref{fig:framework}.
\begin{enumerate}
   \item \textit{Problem and goals.} Consider the medical use case being addressed  (Section~\ref{sec:use-cases}) and any underlying measurements necessary for that use case (Section~\ref{sec:measurements}).
   \item \textit{ML objectives.} From the use case, determine the appropriate ML task and its appropriate objective function (Section~\ref{sec:ml-tasks}).
   \item \textit{Real-world objectives.} In addition to the objective function, consider other real-world objectives pertinent to the use case being addressed (Section~\ref{sec:goals}).
   \item \textit{Elements of interpretability.} From the other real-world objectives, consider the underlying elements which capture these objectives (Section~\ref{sec:elements}).
   \item \textit{Methods.} Given these elements, select the appropriate interpretability method which addresses it (Section~\ref{sec:methods}).
\end{enumerate}
In the remainder of the section, we provide two case studies to illustrate our proposed framework.

\subsection{Case Study 1: Automated, Initial Screening for Diabetic Retinopathy}
\begin{enumerate}
    \item \textit{Problem and goals.}
    Diabetic retinopathy (DR) is a complication of diabetes that can ultimately lead to blindness. 
    Early-stage detection of DR is essential for taking measures to reverse the disease's progression, making it important for people living with diabetes to be screened regularly, with retinal photography to assess for blood vessel damage or retinal lesions being a key imaging modality used for screening~\cite{oh2021earlydetection,beede2020humancentereddr,alyoubi2020diabetic}. 
    However, screening programs often face difficulties due to a shortage of clinical specialists~\cite{beede2020humancentereddr}.
    Therefore, a machine learning solution as an initial triage tool which flags suspicious or high-risk cases and sends them to a clinician for further assessment would reduce clinician burden and enable more widespread screening, which is especially useful in resource-limited environments.
    \item \textit{ML objective.}
    Since this task involves categorization into one of several distinct labels, this is a classification task and the loss function can be a classification loss like cross-entropy.
    \item \textit{Real-world objectives.}
    Beyond classification, as this model will be deployed in a clinical setting, we would demand the ability to troubleshoot and audit the model, and ensure that predictions are fair across patient groups.
    Additionally, user trust of the model is imperative.
    \item \textit{Elements.}
    Of utmost importance amongst the elements is \textbf{localizability} of the lesions or blood vessels which are suspicious.
    Probing a model's ability to localize and recognize the suspicious parts of the image will satisfy the various real-world objectives.
    As this tool will be used as an initial screening tool and the nature of the disease being studied is well understood, the value of \textbf{physical attribution} and \textbf{actionability} is relatively small.
    \item \textit{Methods.}
    Saliency maps may be used to address localizability.
\end{enumerate}

\subsection{Case Study 2: Breast Cancer Detection}

\begin{enumerate}
    \item \textit{Problems and goals.} 
    Early detection of breast cancer is important for increasing the chances of patients’ survival~\cite{nasser2023deep}.
    A large part of clinical diagnosis in this area is concerned with characterizing various features of the tumor from mammograms, including masses, calcification, architectural distortion, asymmetries, and other related signs~\cite{ma2022predicting}. 
    These and other factors have been standardized in the BI-RADS format~\cite{SPAK2017179}.
    Thus, there exists a list of human-recognizable and physically-understood predictive features.
    \item \textit{ML objective.} 
    As the goal of this tool is to make a discrete diagnostic decision, the ML task is classification and the loss function can be a classification loss like cross-entropy.
    \item \textit{Real-world objectives.}
    Similar to Case Study 1, as this model will be deployed in a clinical setting, we would demand the ability to troubleshoot and audit the model, and ensure that predictions are fair across patient groups.
    Additionally, user trust of the model is imperative.
    \item \textit{Elements.}
    Similar to Case Study 1, \textbf{localizability} of the tumors is an important element to consider.
    As this tool will be used as a diagnostic aid by clinicians, it is important to connect to known features that are widely used for breast cancer diagnosis, e.g. as codified by the BI-RADS format.
    Thus, \textbf{visual recognizability} and \textbf{physical attribution} of features are desired.
    \item \textit{Methods.}
    Similar to Case Study 1, saliency maps may be used to address localizability.
    Visual recognizability and physical attribution may be addressed by incorporating feature engineering and radiomics~\cite{karatza2021interpretabilitybreastcancer}.
    The importance of these features can be further interrogated using visualization techniques like ICE and SHAP.
\end{enumerate}

\subsection{Case Study 3: Visual Encoding}
\begin{enumerate}
    \item \textit{Problem and goals.}
    Visual encoding models that predict functional MRI brain responses to visual stimuli are of scientific interest, as they enable quantification of population heterogeneity in brain stimuli-response mapping, which may allow insight into variability in bottom-up neural systems that can in turn be related to individual’s behavior or pathological state~\cite{gu2022personalizedvisualencoding,oota2023deep,wen2017neuralencoding}.
    \item \textit{ML objective.}
    Since this task requires predicting real-valued time-series signals, the task is regression and an appropriate ML loss function might be mean-squared error.
    \item \textit{Real-world objectives.}
    The ultimate objective in this case is scientific discovery. As this model will be employed for uncovering aspects of brain function and how it is linked to individual behavior, we are primarily interested in mechanistic understandings.
    \item \textit{Elements.}
    Important elements include \textbf{model transparency} and \textbf{actionability}, as we are interested in the information and potential scientific understanding which can be uncovered by the model, as well as actionable insights for future experiments.
    \item \textit{Methods.}
    For model transparency, specific architectures can be designed to mimic the brain, like enforcing intermediate layers of the deep network to map to different areas of the visual cortex. 
    These layers may be visualized, for example using t-SNE plots, to understand how the network organizes information in the latent space.
    This can also address actionability, as being able to diagnose issues in some intermediate layers may uncover issues with certain parts of the network.
\end{enumerate}

\section{Discussion}
\label{sec:discussion}
\subsection{Inherent Interpretability of Pixel-level Tasks}
Pixel-level tasks output predictions directly on the image grid, where the level of prediction is localized to the smallest level of granularity.
Often, these outputs represent human-recognizable entities, like region-of-interest labels for segmentation or dense deformation fields for registration.
Thus, it may be seen that such pixel-level tasks exhibit an inherent interpretability because the structures of their predictions happen to be localizable and visually recognizable.
This may explain why there is a relatively fewer number of works focused on interpretability in this context (to the best of our knowledge, there are no survey papers discussing the subject). 

\subsection{Intended Users}
Interpretability is a subjective enterprise, and often times a useful interpretation is in the eye of the beholder.
Thus, keeping in mind intended users is important when considering interpretability methods. 
Methods most useful to clinicians and researchers will differ greatly from methods most useful to model developers and auditors, for example.
Even amongst clinicians, their expertise might vary and will affect what one is looking for in terms of an interpretation.
For example, experienced radiologists might be satisfied with one interpretation, whereas a treating physician might not.

In this work, we have focused on the case where interpretability is sought for deployed systems and where the primary users are clinicians or researchers who may not be experts in ML.
Generally, we argue that localizability, visual recognizability, and physical attribution are the most important elements for these users, as it connects the prediction directly to the images and abstracts away modeling-specific choices.
By contrast, transparency might be more important for model designers who have knowledge of the ML specifics.

\subsection{Faithfulness}
One area of interpretability research not yet mentioned is that of faithfulness of the interpretability method~\cite{goldberg2020towards,dasgupta2022framework,amorim2023faithfulness}.
A faithful interpretable method is one that accurately reflects the underlying predictive model, such that the method is internally coherent~\cite{dasgupta2022framework}. 
Since faithfulness pertains only to the relationship between the explanation and the model, it is relatively removed from connections to the real-world and is thus out-of-scope with respect to this work.
However, practitioners are advised to consider this aspect of interpretability in their own use cases.

\subsection{Limitations of Interpretable \mlmi{}}
While interpretability is a promising direction for the end goal of improving patient care, it can potentially introduce biases or hinder model performance.
Here, we list two potential sources of limitation to consider.

First, interpretable \mlmi{} might induce or contribute to human biases during decision-making processes.
When users' levels of trust in algorithmic decision-making aids are miscalibrated, the result manifests as either automation bias (too much trust) or algorithmic aversion (too much distrust)~\cite{dietvorstAlgorithmAversionPeople, goddardAutomationBiasSystematic2012, parasuramanComplacencyBiasHuman2010}.
False senses of trust can also be induced when models' high-dimensional internal representations are projected into low-dimensional space for human review (e.g., 2-dimensional heatmaps), leading users to the false impression of true understanding while they in fact only have an approximation of ``ersatz'' understanding of the model's inner workings~\cite{babicBewareExplanationsAI2021}.
Because one of the core objectives of interpretable \mlmi{} is to modulate users' trust in ML-powered decision aids, these biases represent pitfalls which must be considered carefully when deploying these models in practice.

Second and relatedly, interpretability in \mlmi{} might run counter to the goal of ultimately augmenting the human user's capabilities.
From this perspective, ``uninterpretability'' might be reflective of a means of performing decision-making orthogonal to the capabilities of humans.
As medicine becomes increasingly complex, the amount of information available to be integrated into the decision-making process may exceed the capacities of the human mind~\cite{obermeyerLostThoughtLimits2017}.
Indeed, constraining ML models to be human-understandable may result in models which overlap with or simply recapitulate patterns already identified by humans.
Leveraging the complementary strengths of human perception and machine perception, including incorporating and processing signals which cannot be easily rationalized by humans, may ultimately be necessarily towards the goal of improving decision-making.

\subsection{Novel Approaches to Interpretability}
While there have been many attempts at interpretability in MLMI, there is still a need for further research and new perspectives.
For example, a nascent approach to interpretability that seeks to reason about the mechanisms by which neural networks arrive at predictions is called mechanistic interpretability~\cite{olah2020zoom}.
Works in this line of research interpret models trained on simple tasks like modular arithmetic and decompose it into individual neurons and circuits which compute different parts necessary to arrive at the answer.
This field has found promise in understanding the inner workings of transformer-based large language models~\cite{elhage2021mathematical}.
While underexplored in the context of medical imaging, mechanistic interpretability is directly aligned with the real-world goal of connecting to and revealing mechanisms and mechanistic understanding. 
We believe works along this vein will be important in the future.

While this work is focused on interpretable ML in medical imaging, we do recognize that similar ideas may be applicable to other domains like law and finance, and hope that our work can inspire others to reason similarly in a domain-specific manner about interpretability.

\section{Conclusion}
We arrived at five elements of interpretability in machine learning for medical imaging through reasoning about real-world goals, tasks, and objectives in medical image analysis and its intersection with machine learning.
These elements are localizability, visual recognizability, physical attribution, model transparency, and actionability. 
Furthermore, we connect our elements to the current literature by summarizing various methods and approaches in interpretable \mlmi{} and attributing each to our elements.
Taken together, these components constitute a framework for interpretability in \mlmi{}, a guide for practitioners on how to reason and utilize interpretability in their own use cases.

\section{Acknowledgements}
We thank Kayhan Batmanghelich for insightful discussions and suggesting the actionability element.
\bibliography{refs}


\begin{IEEEbiography}[{\includegraphics[width=1in,height=1.25in,clip,keepaspectratio]{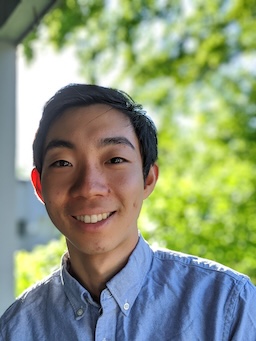}}]{Alan Q. Wang} is a PhD candidate at Cornell University in the School of Electrical and Computer Engineering. He is also affiliated with Cornell Tech and the Department of Radiology at Weill Cornell Medical School. Previously, he obtained his B.S. degree in Computer Engineering at the University of Illinois at Urbana-Champaign (UIUC).

His research interests include machine learning and computer vision, especially applied to medical imaging. 
More specifically, he is interested in bridging the gap between deep learning models and human experts like doctors and clinicians, which leads him to be interested in research in interpretability, explainability, interactivity, and robustness of deep learning models.
\end{IEEEbiography}

\begin{IEEEbiography}[{\includegraphics[width=1in,height=1.25in,clip,keepaspectratio]{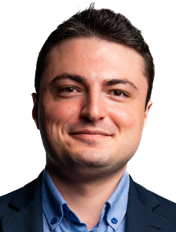}}]{Batuhan K. Karaman} is a PhD candidate at Cornell University in the School of Electrical and Computer Engineering. He is also affiliated with Cornell Tech and the Department of Radiology at Weill Cornell Medicine. Previously, he obtained his MS degree in Electrical and Computer Engineering at Cornell University.

His research interests are in machine learning and biomedical data analysis. Focusing on Alzheimer’s, his current research develops interpretable and explainable deep learning techniques for the early prediction of future cognitive decline toward dementia, and analysis of associated biomarkers.
\end{IEEEbiography}

\begin{IEEEbiography}[{\includegraphics[width=1in,height=1.25in,clip,keepaspectratio]{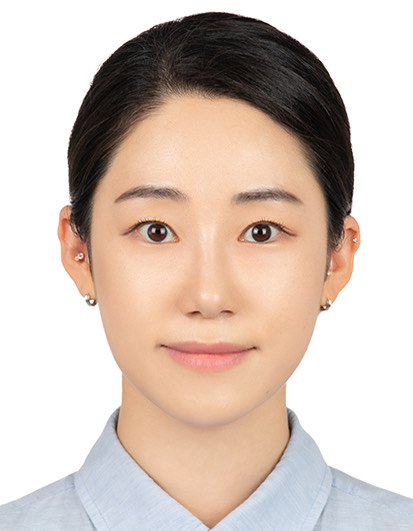}}]{Heejong Kim} is a postdoctoral researcher in Radiology at Weill Cornell Medicine. She completed her Ph.D. in computer science at New York University, New York, NY. Her research interest lies in discovering machine learning techniques and their application to biomedical images. 
\end{IEEEbiography}

\begin{IEEEbiography}[{\includegraphics[width=1in,clip,keepaspectratio]{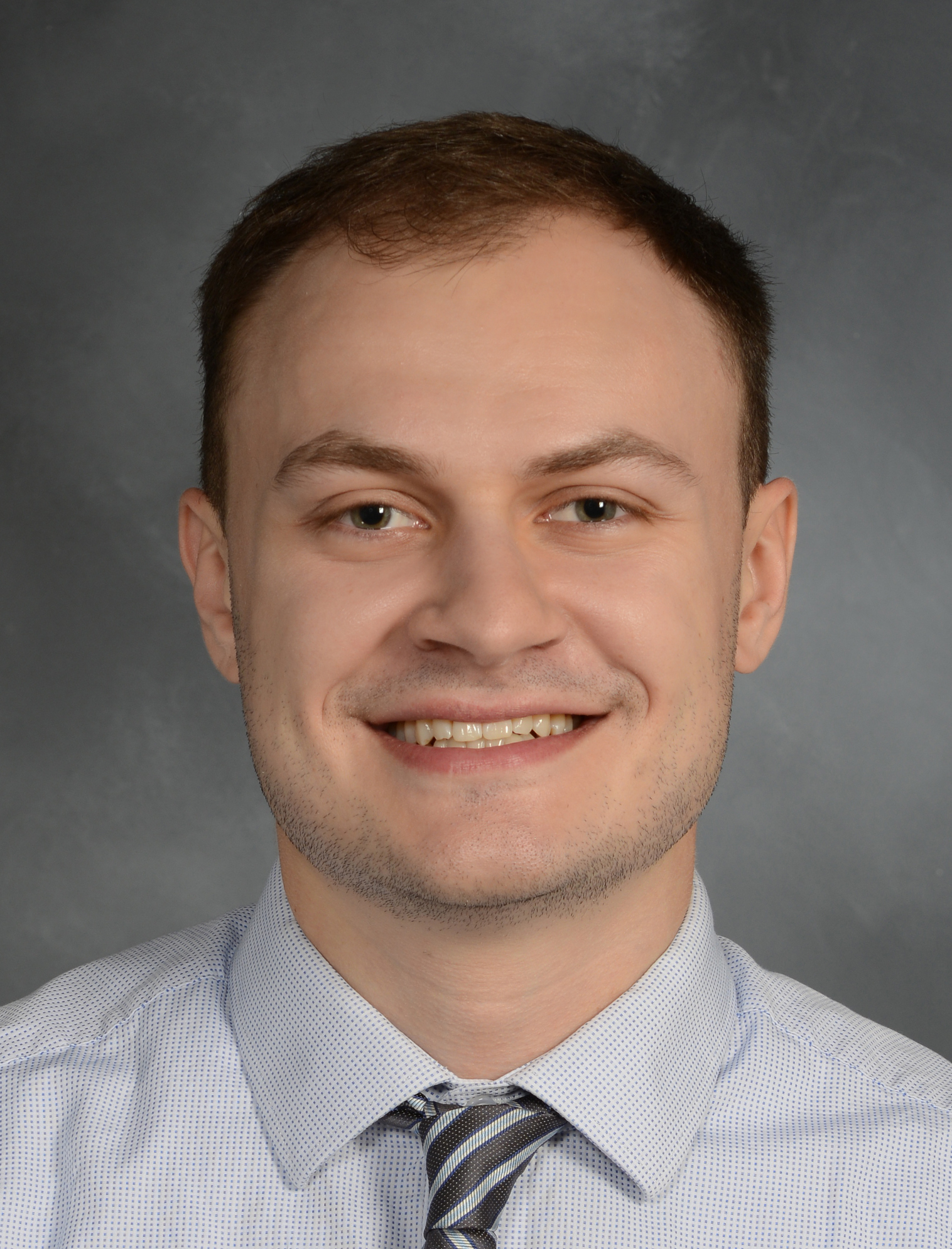}}]{Jacob Rosenthal} is an M.D.-Ph.D. candidate in the Weill Cornell/Rockefeller/Sloan Kettering Tri-Institutional M.D.-Ph.D. Program, New York, NY, USA. Previously, he obtained his M.Sc. degree in health data science from the Harvard T.H. Chan School of Public Health, Boston, MA, USA and his B.A. in biology from Oberlin College, Oberlin, OH, USA.

His research interests lie in the domain of machine learning for medicine, specifically in the development and deployment of machine learning methods to improve accuracy, efficiency, and fairness of clinical workflows.
\end{IEEEbiography}

\begin{IEEEbiography}[{\includegraphics[width=1in,clip,keepaspectratio]{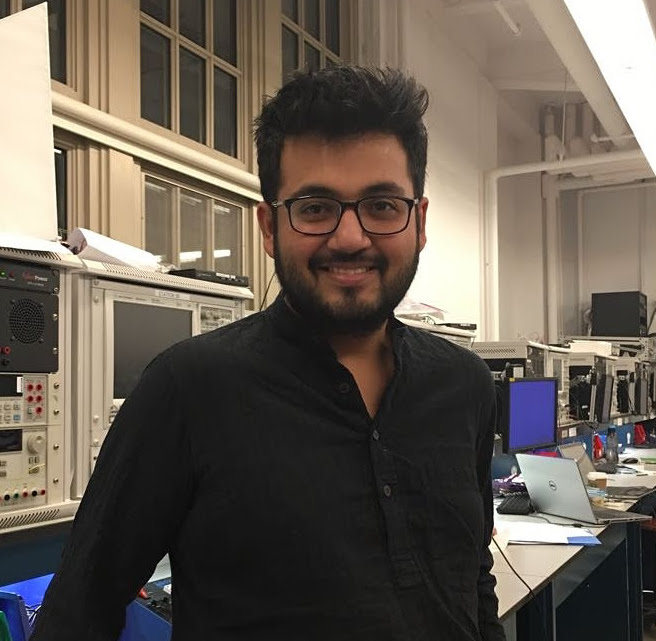}}]{Rachit Saluja} is a PhD candidate at Cornell University in the School of Electrical and Computer Engineering. He is also affiliated with Cornell Tech and the Department of Radiology at Weill Cornell Medical School. Previously, he obtained his M.S. degree in Electrical Engineering from the University of Pennsylvania and obtained his B.E. degree in Electrical and Electronics Engineering from PES Institute of Technology, India. His research interests lie in building Multi-modal AI models for Clinical Radiology and developing robust Radiology AI systems.

\end{IEEEbiography}

\begin{IEEEbiography}[{\includegraphics[width=1in,clip,keepaspectratio]{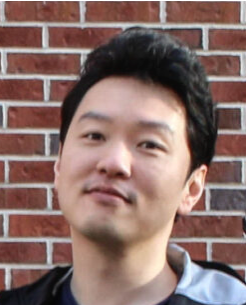}}]{Sean I. Young} is an Instructor at the Martinos Center for Biomedical Imaging, Harvard Medical School, and a Resesarch Affiliate in CSAIL, MIT. Previously, he was a postdoctoral scholar at Stanford University, Stanford, CA. He received his PhD degree in electrical engineering from the University of New South Wales, Sydney, Australia. He received the APRS/IAPR best paper award at DICTA 2018, together with David Taubman.
\end{IEEEbiography}

\begin{IEEEbiography}[{\includegraphics[width=1in,clip,keepaspectratio]{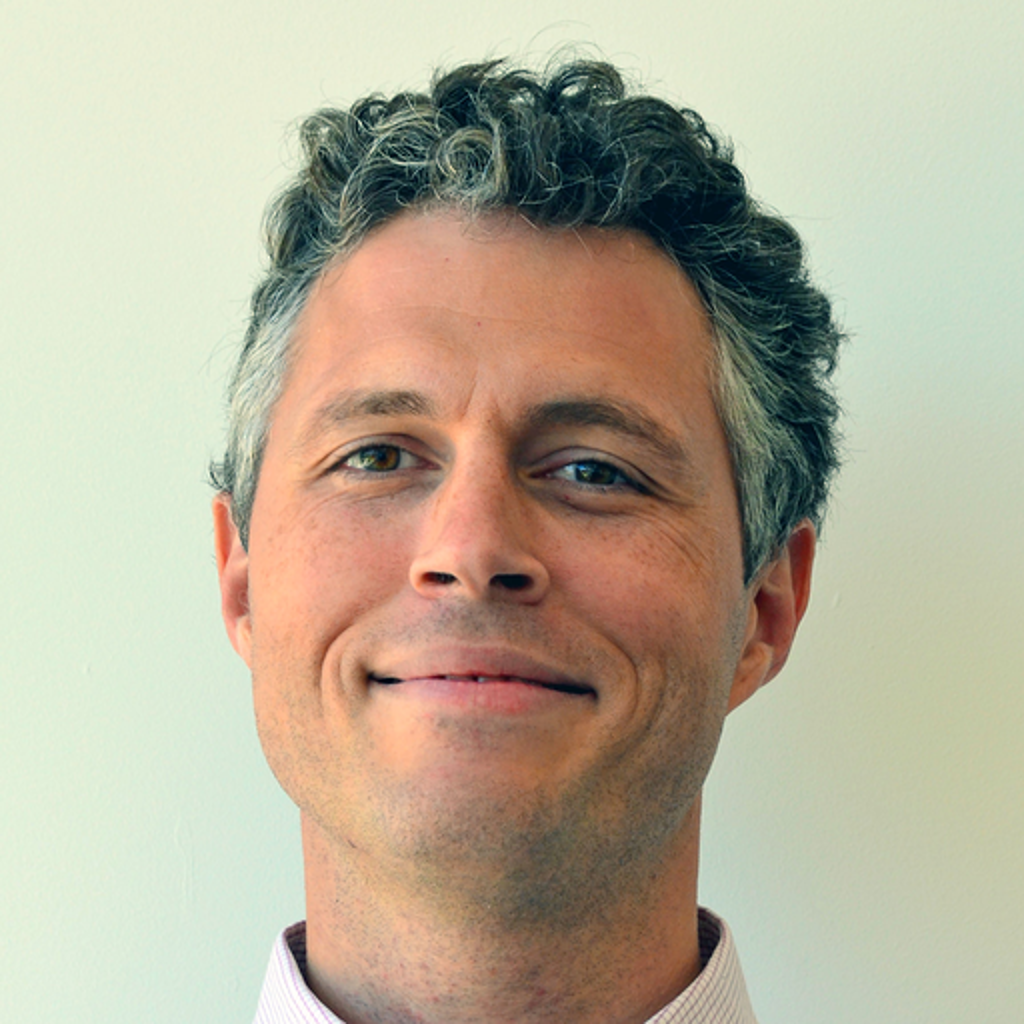}}]{Mert R. Sabuncu} received a PhD degree in Electrical Engineering from Princeton University in 2006, where his dissertation focused on entropy-based approaches to image registration. Mert then moved to the Massachusetts Institute of Technology for a post-doc at the Computer Science and Artificial Intelligence Lab, where he worked on a range of biomedical image analysis problems, including the segmentation of brain MRI scans. After his post-doc at MIT, he spent a few years at the A.A Martinos Center for Biomedical Imaging (Massachusetts General Hospital and Harvard Medical School) as a junior faculty member, where he built a research program on algorithmic tools for integrating genetics and medical imaging. Today, he is a Professor in Electrical and Computer Engineering at Cornell University and Cornell Tech, in New York City. He also holds a faculty appointment in Radiology at Weill Cornell Medicine. His research group develops machine learning based computational tools for biomedical imaging applications. He is a recipient of an NSF CAREER Award (2018) and an NIH Early Career Development Award (2011).
\end{IEEEbiography}
\EOD

\end{document}